\pdfoutput=1

\documentclass[11pt]{article}

\usepackage[]{acl}
\usepackage{mainlp} 
\usetikzlibrary{patterns,patterns.meta}

\title{Different Tastes of Entities: \\ Investigating Human Label Variation in Named Entity Annotations}

\author{Siyao Peng\textsuperscript{\faMountain\kern1pt\faRobot}  \quad 
Zihang Sun\textsuperscript{\faMountain} \quad 
Sebastian Loftus\textsuperscript{\faMountain}  \quad 
Barbara Plank\textsuperscript{\faMountain\kern1pt\faRobot}\\
\textsuperscript{\faMountain} MaiNLP, Center for Information and Language Processing, LMU Munich, Germany \\
\textsuperscript{\faRobot} Munich Center for Machine Learning (MCML), Munich, Germany \\
{\tt 
\{siyaopeng, bplank\}@cis.lmu.de \hspace{0.2em} \{zihang.sun, s.loftus\}@campus.lmu.de }}

\begin{document}
\maketitle

\begin{abstract}
Named Entity Recognition (NER) is a key information extraction task with a long-standing tradition. 
While recent studies address and aim to correct annotation errors via re-labeling efforts, little is known about the sources of human label variation, such as text ambiguity, annotation error, or guideline divergence. This is especially the case for high-quality datasets and beyond English CoNLL03. 
This paper studies disagreements in expert-annotated named entity datasets for three languages: English, Danish, and Bavarian.
We show that text ambiguity and artificial guideline changes are dominant factors for diverse annotations among high-quality revisions. 
We survey student annotations on a subset of difficult entities and substantiate the feasibility and necessity of manifold annotations for understanding named entity ambiguities from a distributional perspective. 
\end{abstract}

\section{Introduction}\label{sec:introduction}

Named Entity Recognition (NER) is a fundamental task in Natural Language Processing (NLP) \citep{yadav_survey_2018}.
The task involves identifying named entities (NEs), such as  \textit{Justin Bieber}, \textit{UNESCO}, and \textit{Costa Rica}, and classifying them into semantic types, \textsc{per}(son), \textsc{org}(anization), and \textsc{loc}(ation), etc. 
Despite recent successes in achieving 93\%+ strict F1 \citep{rucker_cleanconll_2023} on the English CoNLL03 benchmark \citep{tjong_kim_sang_introduction_2003}, recent research has observed that the percentage of noise in the data, particularly in the test partition, is comparable or even exceeding the error rates of state-of-the-art (SOTA) models \citep{wang_crossweigh_2019, reiss_identifying_2020, rucker_cleanconll_2023}.
They each conducted manual corrections or re-annotations, and model performances on their revised versions were higher than on the original. 
However, label variation in NEs, as shown in Table \ref{tab:intro-distribution-student-annotations}, remains an issue and hinders model performance.

Human label variation (i.e., disagreement) refers to linguistically debatable cases where multiple labels are acceptable or appropriate in context \citep{plank_linguistically_2014, jiang_investigating_2022}.
Recent studies that examine and benefit from disagreements among annotators challenge the conventional assumption of a single gold label.
Learning from disagreements provides further insights into label distributions and preferences among human annotators \citep{uma_semeval-2021_2021, plank_problem_2022, fetahu_semeval-2023_2023}.
However, there remains a gap for disagreement analyses on expert-labeled manifold NEs.

{
\setlength{\tabcolsep}{0pt} 
\renewcommand{\arraystretch}{1} 
\begin{table}[t]
\centering
\resizebox{0.49\textwidth}{!}{
\begin{tabular}{cl|ccccc}
\multicolumn{2}{c}{\small{Sentence}}
&  \rotatebox{90}{\tiny{\textsc{per}}}
& \rotatebox{90}{\tiny{\textsc{loc}}}
& \rotatebox{90}{\tiny{\textsc{org}}}
& \rotatebox{90}{\tiny{\textsc{misc}}}
& \rotatebox{90}{\tiny{\textsc{o}}} \\
\hline
\footnotesize{\textit{a.\;}} & \footnotesize{UK bookmakers [William Hill] ... }
& \tikz \fill [black] (0.0, 0.0) rectangle (0.3,0.25); 
& \tikz \fill [black] (0.0, 0.0) rectangle (0.3,0.01); 
& \tikz \fill [black] (0.0, 0.0) rectangle (0.3,0.45); 
& \tikz \fill [black] (0.0, 0.0) rectangle (0.3,0.01); 
& \tikz \fill [black] (0.0, 0.0) rectangle (0.3,0.01); 
\\
\footnotesize{\textit{b.\;}} & \footnotesize{{[ALPINE]} SKIING ...}
& \tikz \fill [black] (0.0, 0.0) rectangle (0.3,0.01); 
& \tikz \fill [black] (0.0, 0.0) rectangle (0.3,0.3); 
& \tikz \fill [black] (0.0, 0.0) rectangle (0.3,0.01); 
& \tikz \fill [black] (0.0, 0.0) rectangle (0.3,0.15); 
& \tikz \fill [black] (0.0, 0.0) rectangle (0.3,0.20); 
\\
\footnotesize{\textit{c.\;}} & \footnotesize{... that there is no [God] .}
& \tikz \fill [black] (0.0, 0.0) rectangle (0.3,0.50); 
& \tikz \fill [black] (0.0, 0.0) rectangle (0.3,0.01); 
& \tikz \fill [black] (0.0, 0.0) rectangle (0.3,0.01); 
& \tikz \fill [black] (0.0, 0.0) rectangle (0.3,0.20); 
& \tikz \fill [black] (0.0, 0.0) rectangle (0.3,0.01); 

\end{tabular}
}
\caption{Distribution of qualified student annotations on disagreed named entities in CoNLL03.}
\label{tab:intro-distribution-student-annotations}
\end{table}
}

This paper presents quantitative and qualitative analyses of annotators' disagreements on labeling NEs in three Germanic variants: English, Danish, and Bavarian, in which multiple annotation efforts exist on the same documents. 
Unlike earlier studies that look at crowd-sourced data of unreliable quality \citep{rodrigues_sequence_2014, lu_are_2023},
we examine disagreements among expert annotations that went through iterations of published revisions and contrast them with the usual setting of independent annotators. 
\S\ref{sec:related-work} presents related work in disagreements and \S\ref{sec:datasets_preprocessing} demonstrates our setups.
We analyze entity and label disagreements in \S\ref{sec:entity-level-disagreements},  sources of disagreements in \S\ref{sec:source-disagreement}, and a student-surveyed annotation study in \S\ref{sec:survey-student-annotations}.
\S \ref{sec:conclusion} summarizes our work.
We release our annotations and analyses on Github.\footnote{\url{https://github.com/mainlp/NER-disagreements/}}

\section{Related Work}\label{sec:related-work}

Despite disagreements between human judgments in subjective tasks \citep{prabhakaran_releasing_2021, davani_dealing_2022, fetahu_semeval-2023_2023, leonardelli_semeval-2023_2023}, 
annotation variation studies in NLP are recently on the rise \citep{uma_semeval-2021_2021, plank_problem_2022, fetahu_semeval-2023_2023}.
These include
part-of-speech tagging \citep{plank_linguistically_2014},
anaphora and pronoun resolution \citep{poesio_reliability_2005, poesio_crowdsourced_2019, haber_classication_2020},
discourse relation labeling \citep{marchal_establishing_2022, pyatkin_design_2023},
word sense disambiguation \citep{passonneau_multiplicity_2012, navigli_semeval-2013_2013, martinez_alonso_predicting_2015},
natural language inference \citep{nie_what_2020, jiang_investigating_2022, liu_were_2023},
question answering \citep{min_ambigqa_2020, ferracane_did_2021},
to name a few.

In NER, 
\citet{rodrigues_sequence_2014} crowd-sourced problematic annotations from 47 Turkers on CoNLL03, scoring F1 of 17.60\% the lowest and $\sim$60\% on average against CoNLL03 annotations, 
considerably under-performing the 90\%+ inter-annotator agreement among expert annotators and SOTA model performances \citep{lu_are_2023}. 
Recently, \citet{rucker_cleanconll_2023} brought forward the newest CoNLL03 correction and thoroughly compared it with previous versions \citep{tjong_kim_sang_introduction_2003, wang_crossweigh_2019, reiss_identifying_2020}. 
However, many corrections are due to project-dependent guideline alternations and 2.34\% of entities remain unresolved due to ambiguities.
Thus, an onlooker assessment of NE disagreements and label variations is missing, particularly for expert annotations.

\section{Datasets \& Preprocessing}\label{sec:datasets_preprocessing}

We analyze label variations in CoNLL03-styled \textsc{per/loc/org/misc} NE annotations in three Germanic languages: English, Danish, and Bavarian (a Germanic dialect without standard orthography), where multiple annotation efforts on the same text documents are (or will be) available. 
Since the English CoNLL03 \citep{tjong_kim_sang_introduction_2003} and Danish DDT \citep{plank_neural_2019} texts underwent iteration(s) of re-annotations or corrections by subsequent scholars, we conduct a diachronic comparison of the revisions for English and Danish.
We also analyze disagreements on an in-house NE dataset for Bavarian German to distinguish disagreements among full-fledged corpora from independent unadjudicated annotations.

\paragraph{English}
The seminal English CoNLL03 dataset (henceforth \texttt{original}, \citealt{tjong_kim_sang_introduction_2003}) presents the renowned NLP task to label flat and named entity spans into four major semantic types (\textsc{per, loc, org, misc}) using (B)IO-encoding.
The dataset includes 14.04K, 3.25K, and 3.45K sentences in its train, dev, and test partitions sourced from Reuters News between 1996-1997.
Despite achieving 93\%+ F1 score of the best systems on \texttt{original}, 
CoNLL03 annotations underwent several revisions \citep{
wang_crossweigh_2019,  
reiss_identifying_2020,  rucker_cleanconll_2023}.

\citet{wang_crossweigh_2019} (\texttt{conllpp}) manually corrected 186 (5.38\%) test sentences.
\citet{reiss_identifying_2020} (\texttt{reiss}) used a semi-automatic approach to flag a larger quantity of error-prone labels (3.18K) in the entire dataset, and manually corrected 1.32K, including 421 in the test, as well as fixing tokenization and sentence splitting. 
They categorize these errors into six types: \texttt{Tag, Span, Both, Wrong, Sentence}, and \texttt{Token}.
\citet{rucker_cleanconll_2023} (\texttt{clean}) present the most comprehensive relabeling effort by correcting 7.0\%  of all labels and adding a novel layer for entity linking. 
Though 5\%+ of annotation errors were fixed compared to \texttt{original}, 2.34\% of entities in \texttt{clean} remain ambiguous.

To establish fair comparisons, we manually align tokenization in the test partitions of \texttt{original, conllpp, reiss}, and \texttt{clean}.
These include removing redundant line breaks,
splitting hyphenized compounds,
etc.
Our alignment results in 46,738 test tokens across the four versions
and 5,629, 5,683, 5,636, 5,725 annotated entities respectively.

\paragraph{Danish}
\citet{plank_neural_2019} annotates NEs on the dev and test partitions of the Danish Universal Dependencies (DDT, \citealt{johannsen2015universal}).
\citet{plank_dan_2020} (\texttt{plank})
revise annotations, expand to more data and genres,  
and add -part/deriv suffixed labels and second-level nesting.
\citet{hvingelby_dane_2020} (\texttt{hvingelby}) re-annotate the dev and test sets of \citet{plank_neural_2019} by adding POS-marked proper nouns as NEs, resulting in $\sim$0.75 and $\sim$3.0 times more \textsc{org} and \textsc{misc} NEs, such as nationalities and derived adjectives.
We focus on the test partition (10,023 tokens) and compare \texttt{hvingelby} to the more recent \texttt{plank}, removing nesting and -part/deriv entities for cross-lingual analogy, leading to 564 and 531 NEs in \texttt{hvingelby} and  \texttt{plank}.

\paragraph{Bavarian}
We additionally analyze the test partition of an in-house Bavarian NE dataset with $\sim$12K tokens and $\sim$400 entities on Wikipedia and Twitter (X) annotated in 2023.
Compared to the more established and iteratively revised English and Danish datasets, 
our Bavarian corpus represents the more common scenario of disagreements between two independent and unadjudicated annotations.

\section{Entity-level Disagreements}\label{sec:entity-level-disagreements}

Given our manually aligned tokenization across datasets, we modify \citet{reiss_identifying_2020}'s six error types into four entity-level disagreement types: 
\begin{itemize}[itemsep=0pt]
\item \texttt{Tag}: same span selection, but different assigned tags, e.g., \textit{[a b]\textsc{loc}} vs. \textit{[a b]\textsc{org}};
\item \texttt{Span}: different overlapping spans but the same tag, e.g., \textit{[a b]\textsc{loc}} vs. 
 \textit{[a]\textsc{loc} b};
\item \texttt{Both}: overlapping spans with different tags, e.g., \textit{[a b]\textsc{loc}} vs. \textit{[a]\textsc{org} b};
\item \texttt{Missing}: one annotator misses the entity completely, e.g., \textit{[a b]\textsc{loc}} vs. \textit{a b}.
\end{itemize}

\begin{figure}[t]
\centering
\includegraphics[width=0.50\textwidth]{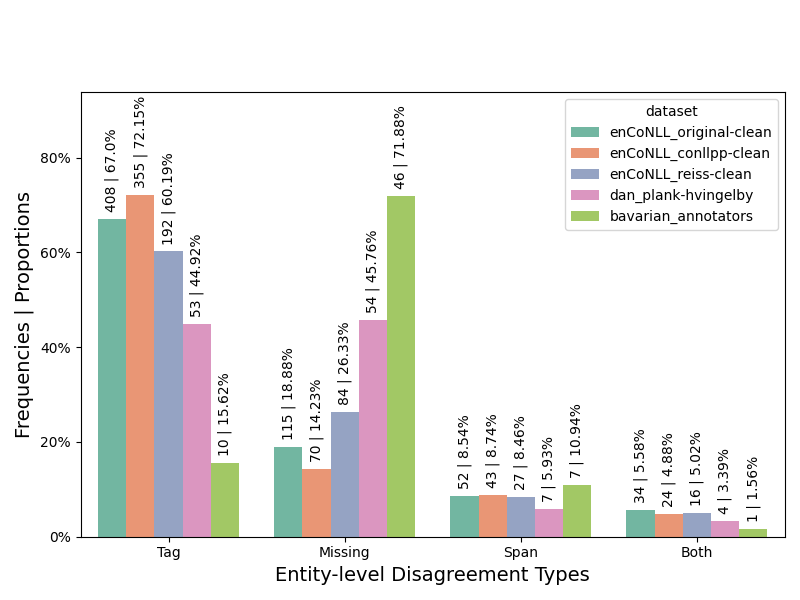}
\caption{Proportions of entity-level disagreements in English \texttt{original-clean, conllpp-clean, reiss-clean}, Danish \texttt{plank-hvingelby}, and Bavarian.}
\label{fig:beyond-label-disagreements}
\end{figure}

\noindent
Figure \ref{fig:beyond-label-disagreements} presents the frequencies and proportions of entity-level disagreements in five paired comparisons: English \texttt{original-clean, conllpp-clean, reiss-clean}, Danish \texttt{plank-hvingelby}, and between two Bavarian annotators.
\texttt{Tag} disagreements contribute to most cases among repeatedly developed English corpora. 
On the other hand, Danish and Bavarian contain more \texttt{Missing} disagreements. 
Nevertheless, combining \texttt{Tag} and \texttt{Missing} accounts for 85\%+ of disagreements in all comparisons across three languages. That is, entity tagging remains a bigger issue compared to span selection.

\begin{figure}[t]
    \centering
    \includegraphics[width=0.50\textwidth]{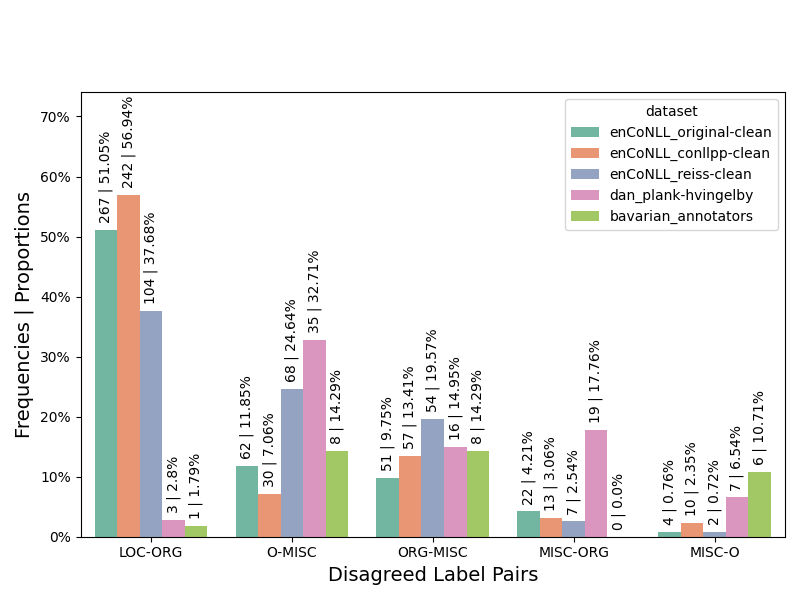}
    \caption{Proportions of top 5 label pairs in \texttt{Tag} and \texttt{Missing} disagreements in English, Danish, and Bavarian.}
    \label{fig:disagreed-label-pairs-top5}
\end{figure}

\texttt{Tag} and \texttt{Missing} disagreements are comparable in that both concern tagging the same entity span with different labels: the former with two different entity types (i.e., two non-\textsc{o} labels), and the latter with one entity type (a non-\textsc{o} label) and an \textsc{o}. 
Figure \ref{fig:disagreed-label-pairs-top5} displays the proportions of the top 5 disagreed label pairs in \texttt{Tag} and \texttt{Missing} disagreements across the five comparison scenarios (see Appendix \ref{sec:appendix-proportion-disagreed-pair} for a full list of label pairs).
\textsc{loc-org, o-misc} and \textsc{org-misc} are the most frequently disagreed label pairs in English comparisons, totaling 70\%+ label disagreements. 
On the other hand, most (80\%+) of Danish label disagreements concern \textsc{misc}, whereas \textsc{o}-related (i.e., \texttt{Missing}) disagreements donate the majority (70\%+) to Bavarian.
To understand which factors trigger these label disagreements, \S\ref{sec:source-disagreement} qualitatively analyzes the sources of human label variations in three languages.

\section{Sources of Disagreements}\label{sec:source-disagreement}

\paragraph{Taxonomy}
We attribute NE label variations to three sources \citep{aroyo_truth_2015, jiang_investigating_2022}:
1) \textit{text ambiguity} for uncertainties in the sentence meaning, 
2) \textit{guideline update} where NE type definitions vary across different guideline versions,
and 3) \textit{annotator error}. 
\textit{Text ambiguity} could be caused by different interpretations with or without enough context that hinders pinpointing a definitive reference.
\textit{Guideline update} occurs when one annotation version is incoherent with another guideline. 
This is dominant in our analyses since
annotation projects consist of iterations of guidelines and annotation revisions.
For instance, whether proper noun-derived adjectives, e.g., \textit{[ALPINE]} in Table \ref{tab:intro-distribution-student-annotations}, should be \textsc{loc}, \textsc{misc}, or not an entity (i.e., \textsc{o});
whether polysemous \textsc{loc/org} entities are labeled \textsc{loc} or \textsc{org} depending on context, or always as \textsc{misc}.
The last category, \textit{annotator error}, refers to annotations that differ from a single deterministic ground truth.
Closer inspections could fix annotators' attention slip errors, whereas special cultural knowledge is needed for resolving knowledge gap disagreements.
We manually annotate a small sample of disagreements in three languages using these source categories to separate guideline changes and textual ambiguities from annotators' mistakes. 

\paragraph{Setup} For English, we sample 200 disagreed test entities between the \texttt{original} and the most recent \texttt{clean} annotation. 
Since the Danish \texttt{plank- hvingelby} comparisons and the Bavarian double annotations have much smaller test sets, we sample all test disagreements in the two languages,  118 entities in Danish and 64 in Bavarian.
Each language sample is assessed by one computational linguist who speaks that language. 
Table \ref{tab:disagreement_sources_proportions} presents the source of disagreement results.

Additionally, we measure inter-annotator agreement (IAA) on source classes between two assessors\footnote{We use ``assessors'' to refer to our source of disagreement coders and differentiate from ``annotators'' of the NE datasets.} on 50 ambiguous English \texttt{original-clean} test entities and achieve 61.73\% Cohen's kappa. 
Assessors find the hardest differentiating whether the lack of contextual information resulted from annotators' personal knowledge backgrounds (\textit{annotator error)} or the settings behind text segments (\textit{text ambiguity}). 
Even though surrounding sentences are provided, NE annotators tend to focus on the nearer context for NE tagging.

\begin{table}[t]
\centering
\resizebox{0.49\textwidth}{!}{
\begin{tabular}{l|rrr}
\hline
Source types
& \multicolumn{1}{c}{English}
& \multicolumn{1}{c}{Danish} 
& \multicolumn{1}{c}{Bavarian} 
\\
\hline
text ambiguity 
& 19 $|$ \; \; 9.5\% 
&  7 $|$ \;  \; 6.0\% 
& 10 $|$ \; 15.6\% \\
\hline
guideline update 
& 160 $|$ \;  80.0\% 
&  62 $|$ \; 52.5\%  
&  11 $|$ \; 17.2\% \\
\hline
annotator error 
& 21 $|$ \; 10.5\% 
& 49 $|$ \; 41.5\%
& 43 $|$ \; 67.2\% \\
\hline
Total & 
200 $|$ 100.0\% 
& 118 $|$ 100.0\% 
& 64 $|$ 100.0\%
\end{tabular}
}
\caption{Sources of label disagreements and their distributions in English, Danish, and Bavarian samples.}
\label{tab:disagreement_sources_proportions}
\end{table}

\paragraph{English}

In the \texttt{original-clean} comparison, 
most (80.0\%) of disagreements stem from differences in \textit{guideline update}. To disambiguate inconsistent cases in \texttt{original}, \texttt{clean} updated the guideline to be less context-dependent: 
1) \textsc{org} instead of \textsc{loc} for national sports teams as well as public facilities,
even for \textit{the flight to [Atlanta]\textsc{org}}; 
2) \textsc{misc} is used for more abstract institutions
and adjectival affiliations
e.g., \textit{[Czech]\textsc{misc} politics};
3) instead of further correcting tokenizations and splitting hyphenated compounds, they assign labels that are relevant to part of the compound to the entirety, e.g., \textit{[German-born]\textsc{misc}}.
Aside from \textit{guideline update}, ambiguities occur for religious deities, such as whether \textit{[Allah]} or \textit{[God]} should be \textsc{per, misc} or \textsc{o} (see Table \ref{tab:intro-distribution-student-annotations}).
Previous automatic conversions from IO-encodings in \texttt{original} to BIO in \texttt{clean} also caused disagreements
since it is hard to tell apart if a sequence of \textsc{i-}tags is one entity or multiple,
e.g., \textit{[Spanish]\textsc{misc} [Super Cup]\textsc{misc}} or  \textit{[Spanish Super Cup]\textsc{misc}}.

\paragraph{Danish}

Akin to the English analysis, we found that large parts (52.5\%) of the ambiguous cases in Danish stem from \textit{guideline updates}, e.g.,  frequently mentioned ferry routes are labeled \textsc{loc} in \texttt{hvingelby} but \textsc{misc} in \texttt{plank}. 
Besides, we found 41.5\% of disagreements are \textit{annotator errors}, and the majority are \textsc{org-misc} disagreements and concern a single hyphen-joint token with two sports clubs, e.g., \textit{[Vejle-Ikast]}. 
This points out a disadvantage of the current cross-lingual comparable analysis --- compounding morphology prevails in Danish and Bavarian, and removing -part/deriv labels leads to information loss. 

\paragraph{Bavarian}
We present the less developed but more common scenario of disagreements between two unadjudicated annotations in Bavarian.
Though achieving 85\%+ Span IAA, 
\textit{annotator error} (67.2\%) remains the highest source of disagreements.
Apart from local entities, e.g., \textit{[Feucht]}\textsubscript{loc} (a small town in Bavaria),
that require geographical knowledge or detailed search,  many of these \textit{annotator errors} classified based on the Bavarian guideline are indeed acceptable under certain versions of the English CoNLL guidelines.
For example, when \textit{[Edeka]} (a supermarket chain) functions as a destination, the disagreement between \textsc{loc-org} is classified as \textit{annotator error} in Bavarian, but would rather be a \textit{guideline update} in English.

\section{Surveying Student Annotations}\label{sec:survey-student-annotations}

Though NE guidelines can be meticulously different from each other, the underlying concepts of 
\textsc{per, loc, org} are cognitively straightforward. 
To inspect the distribution of multiple interpretations, we follow \citet{liu_were_2023} to survey annotations from 27 bachelor and master students in computational linguistics at LMU Munich.
We gave them a 7-minute introduction to NEs, 
walked through the CoNLL03 guideline\footnote{\href{https://www.cnts.ua.ac.be/conll2003/ner/annotation.txt}{\texttt{www.cnts.ua.ac.be/conll2003/ner/annotation.txt}}}, 
and showed some examples of type ambiguities in NE annotations.
Students were instructed in the classroom to annotate entity types in English and Bavarian selected from difficult examples in \S\ref{sec:source-disagreement}.\footnote{Students acknowledge that their annotations could be used for research purposes.} 
We further sample 10 representative English CoNLL entities for the qualitative evaluation below.\footnote{The full English and Bavarian student-surveyed annotations are available on GitHub.}
To ensure the quality of student-surveyed annotations, we only keep an annotation if 80\%+ of entity labels match any of the four CoNLL annotations.
Table \ref{tab:intro-distribution-student-annotations} demonstrates the distribution of 14 qualified student annotations on three examples (see Appendix \ref{sec:appendix-survey-annotations} for the ten representative English CoNLL entities).

Results demonstrate that label variation across annotation projects are also prevalent in the student-surveyed annotations. 
On one side, even with a brief training, students were able to disambiguate the contextual interpretations between [the away team]\textsc{org} and [the home team]\textsc{loc} in \textit{[LA CLIPPERS]\textsc{org} AT [NEW YORK]\textsc{loc}}.
Our participants also recognize the collectiveness of \textit{[White House]\textsc{org}}, \textit{[Australia]\textsc{org}}, etc.,
and the fixedness of \textit{[EST]\textsc{misc}} (Eastern Standard Time). 
On the other hand, knowledge gap or insufficient context contribute to the high variance of \textit{[William Hill]}, whether it refers to [the businessman]\textsc{per} or [the gambling bookstore he created]\textsc{org}. 
Annotators also diverge in marginal cases: whether \textit{[God]} is \textsc{per} or \textsc{misc} and whether nominal derivatives \textit{ALPINE} and \textit{Fascist} are NEs.  

\section{Conclusion}\label{sec:conclusion}

This paper examines named entity disagreements across expert annotations and contrasts them with the more common setting of individual annotations. 
We demonstrate that human label variation, e.g., \textsc{loc-org} and \textsc{org-misc}, contribute to most English, Danish, and Bavarian disagreements.
We also discover that \textit{guideline updates} and \textit{text ambiguities} are leading sources of disagreements in established English and German datasets, 
whereas \textit{annotator errors} remain the dominant cause for the new Bavarian corpus.
Lastly, we survey student annotations and encourage more researchers to explore NE label variations to narrow the gap to model performance. 

Though modeling NER from label variation is out of the scope of this paper, we embrace the prospect of learning from disagreements \citep{uma_learning_2021}.
Particularly, we look forward to conducting annotations on a much larger scale in terms of both the number of participants and annotated instances to provide more statistically meaningful NE distributions for NER models.
Future work also includes separating valid label variations from true annotation mistakes by leveraging   
Automatic Error Detection (AED) methods \citep{klie-aed-2023, weber_activeaed_2023}. 
We hope tackling NER through label variations can remedy the conflicts among versions of annotation guidelines. 


\section*{Acknowledgements}
We would like to thank Verena Blaschke for giving feedback on earlier drafts of this paper.  
This project is supported by ERC Consolidator Grant DIALECT 101043235.

\bibliography{zotero, mypaper}

\newpage 
\appendix

\section{Proportions of Disagreed Label Pairs}\label{sec:appendix-proportion-disagreed-pair}

\begin{figure}[t!bh]
\centering
\includegraphics[width=0.50\textwidth]{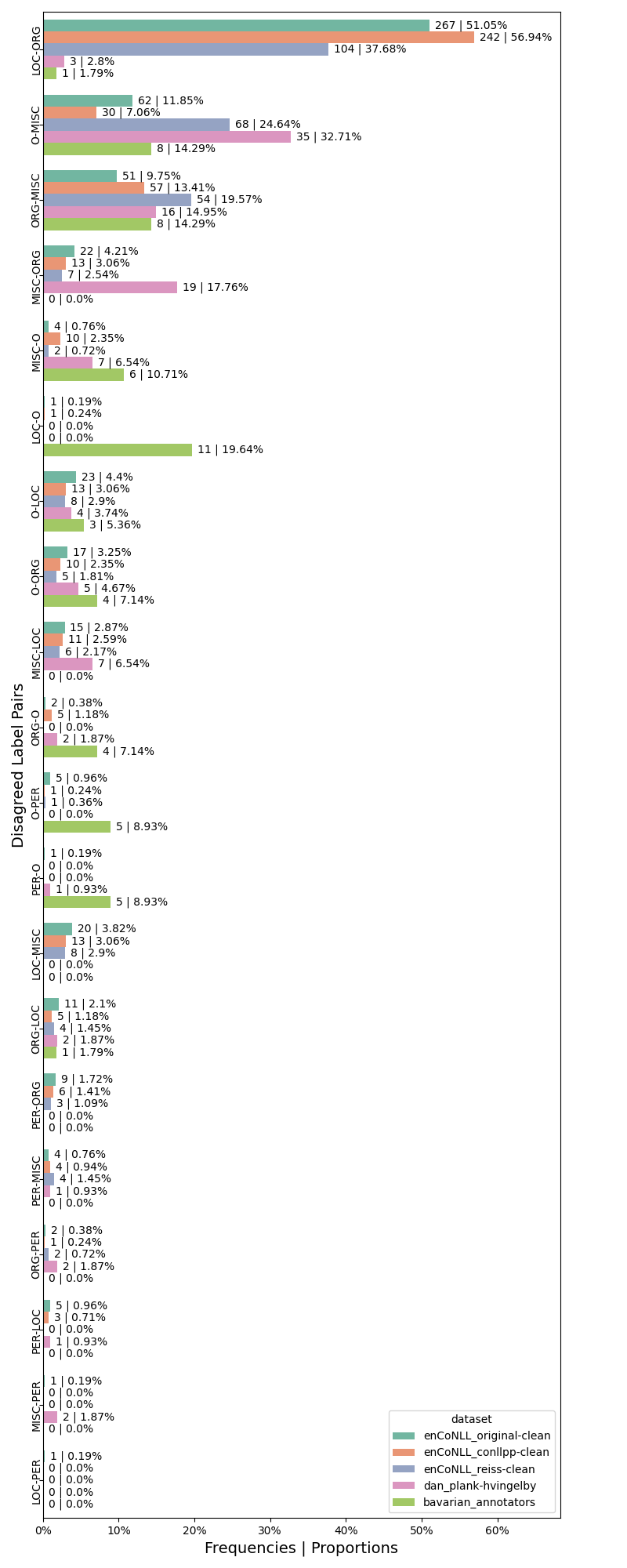}
\caption{Proportions of label pairs (full) in \texttt{Tag} and \texttt{Missing} disagreements in English, Danish, and Bavarian.}
\label{fig:disagreed-label-pairs-full}
\end{figure}

\section{Student Surveyed NE Annotations}\label{sec:appendix-survey-annotations}
\begin{table*}[ht]
\centering
\resizebox{0.99\textwidth}{!}{
\renewcommand{\arraystretch}{1.0}
\begin{tabular}{c|cccccc}
 Sentence & \textsc{per} & \textsc{loc} & \textsc{org} & \textsc{misc} & \textsc{o} & \textit{abstained} \\
\hline
{\textit{[ALPINE]}} \textit{SKIING}
&  
& \begin{tabular}[t]{@{}c@{}} 6 \\
\texttt{clean}
\end{tabular}
&  & 3
& \begin{tabular}[t]{@{}c@{}} 4 \\
\texttt{original} \\
\texttt{conllpp} \\
\texttt{reiss} \\
\end{tabular}
& 1
 \\
{\textit{[LA CLIPPERS]}} \textit{AT NEW YORK}
&  &  
&  \begin{tabular}[t]{@{}c@{}} 13 \\
\texttt{original} \\
\texttt{conllpp} \\
\texttt{reiss} \\
\texttt{clean}
\end{tabular}
&  & & 1
 \\
 \textit{LA CLIPPERS AT [NEW YORK]}
& 
& \begin{tabular}[t]{@{}c@{}} 14 \\
\texttt{original} \\
\texttt{conllpp} \\
\texttt{reiss} \\
\end{tabular}
& \begin{tabular}[t]{@{}c@{}}  0 \\
\texttt{clean}
\end{tabular}
&    & & 
 \\
 \begin{tabular}[c]{@{}c@{}}
 {\textit{[White House]}} \textit{spokesman Mike McCurry said Clinton plans} \\ \textit{to have regular news conferences during his second term .}
 \end{tabular}
&  
& \begin{tabular}[t]{@{}c@{}} 2 \\
\texttt{original} \\
\texttt{conllpp} \\
\texttt{reiss}
\end{tabular}
& \begin{tabular}[t]{@{}c@{}} 11 \\
\texttt{clean}
\end{tabular}
& 1 &  & 
 \\
 \begin{tabular}[c]{@{}c@{}}
\textit{UK bookmakers [William Hill]}\footnote{\textit{William Hill} is a British gambling company} \textit{said on Friday they} \\ \textit{have lengthened the odds of a Conservative victory .}
 \end{tabular}
& \begin{tabular}[t]{@{}c@{}} 5 \\
\texttt{original} \\ 
\texttt{conllpp} \\ 
\texttt{reiss}
\end{tabular}
& 
& \begin{tabular}[t]{@{}c@{}} 9  \\
\texttt{clean}
\end{tabular}
&  &  & 
 \\
\begin{tabular}[c]{@{}c@{}}
\textit{The man who kicked [Australia] to defeat with a last-ditch} \\ \textit{drop-goal in the World Cup quarter-final in Cape Town .}
\end{tabular}
& 
&\begin{tabular}[t]{@{}c@{}}  5  \\
\texttt{original} \\
\texttt{conllpp} \\
\texttt{reiss}
 \end{tabular}
& \begin{tabular}[t]{@{}c@{}} 9 \\
\texttt{clean}
\end{tabular} 
&   &  & 
 \\
\begin{tabular}[c]{@{}c@{}}
\textit{The years I spent as (soccer team) manager of} \\ {\textit{the [Republic of Ireland]}} \textit{were the best years of my life .}
\end{tabular} 
& 
& \begin{tabular}[t]{@{}c@{}}  4 \\
\texttt{original} \\ 
\texttt{conllpp} \\ 
\texttt{reiss}
\end{tabular}
&  \begin{tabular}[t]{@{}c@{}} 9 \\
\texttt{clean}
\end{tabular}
& 1 &  & 
 \\
\textit{I bear witness that there is no [God] .}
& \begin{tabular}[t]{@{}c@{}} 10 \\
\texttt{original} \\ 
\texttt{conllpp} \\ 
\texttt{reiss}
\end{tabular}
&  & 
& \begin{tabular}[t]{@{}c@{}} 4 \\
\texttt{clean}
\end{tabular}
& 
& 
 \\
 \begin{tabular}[c]{@{}c@{}}
\textit{The granddaughter of Italy’s [Fascist]}\footnote{\textit{Fascism} is a far-right, authoritarian, ultranationalist political ideology and movement} \textit{dictator Benito Mussolini}
\end{tabular}
&  &  & 3 
& \begin{tabular}[t]{@{}c@{}} 3  \\
\texttt{clean}
\end{tabular}
& \begin{tabular}[t]{@{}c@{}} 8 \\ 
\texttt{original} \\ 
\texttt{conllpp} \\ 
\texttt{reiss}
\end{tabular} 
& 
 \\
\begin{tabular}[c]{@{}c@{}}
\textit{at about 3 A.M. local time / 1:30 A.M. [EST]}
\end{tabular}
&  &  &  
& \begin{tabular}[t]{@{}c@{}} 10 \\
\texttt{clean}
\end{tabular} 
& \begin{tabular}[t]{@{}c@{}} 2 \\
\texttt{original} \\ 
\texttt{conllpp} \\ 
\texttt{reiss}
\end{tabular} 
& 2
 \\ 
\end{tabular}
}
\caption{14 classroom surveyed and qualified annotations on difficult disagreement cases in CoNLL03 test.}
\label{tab:crowd_source_annotations_full}
\end{table*}


\end{document}